\documentclass[journal,twoside]{IEEEtran}

\usepackage{cite}

\ifCLASSINFOpdf
  \usepackage[pdftex]{graphicx}
\else
\fi

\usepackage{amsmath}

\usepackage{array}

\usepackage{amssymb} %
\usepackage{booktabs} %

\renewcommand{\arraystretch}{1.2} %

\usepackage{lipsum}
\usepackage{xcolor}

\begin{document}

\title{Interaction Relational Network for\\ Mutual Action Recognition}

\author{Mauricio~Perez,~\IEEEmembership{Member,~IEEE,}
        Jun~Liu,
        and~Alex~C.~Kot,~\IEEEmembership{Fellow,~IEEE}%
\thanks{M. Perez and A. C. Kot are with the Rapid-Rich Object Search Lab, School of Electrical and Electronic Engineering, Nanyang Technological University, Singapore, 639798. E-mail: \{mauricio001, eackot\}@ntu.edu.sg}
\thanks{J. Liu is with ISTD Pillar, Singapore University of Technology and Design, Singapore. E-mail: {jun\_liu}@sutd.edu.sg}
\thanks{Corresponding author: Jun Liu}
}

\maketitle

\begin{abstract}

Person-person mutual action recognition (also referred to as interaction recognition) is an important research branch of human activity analysis. 
Current solutions in the field -- mainly dominated by CNNs, GCNs and LSTMs -- often consist of complicated architectures and mechanisms to embed the relationships between the two persons on the architecture itself, to ensure the interaction patterns can be properly learned.
Our main contribution with this work is by proposing a simpler yet very powerful architecture, named Interaction Relational Network, which utilizes minimal prior knowledge about the structure of the human body. 
We drive the network to identify by itself how to relate the body parts from the individuals interacting.
In order to better represent the interaction, we define two different relationships, leading to specialized architectures and models for each. These multiple relationship models will then be fused into a single and special architecture, in order to leverage both streams of information for further enhancing the relational reasoning capability.
Furthermore we define important structured pair-wise operations to extract meaningful extra information from each pair of joints -- distance and motion. 
Ultimately, with the coupling of an LSTM, our IRN is capable of paramount sequential relational reasoning.
These important extensions we made to our network can also be valuable to other problems that require sophisticated relational reasoning.
Our solution is able to achieve state-of-the-art performance on the traditional interaction recognition datasets SBU and UT, and also on the mutual actions from the large-scale dataset NTU RGB+D. Furthermore, it obtains competitive performance in the NTU RGB+D 120 dataset interactions subset.

\end{abstract}


 \ifCLASSOPTIONpeerreview
 \begin{center} \bfseries EDICS Category: 9-DLMA \end{center}
 \fi
\IEEEpeerreviewmaketitle

\section{Introduction} %
\label{sec:intro}

\IEEEPARstart{R}{ecognition} of human interaction (mutual actions) in videos is an important computer vision task, which can help us to develop solutions for a range of applications, such as surveillance, robotics, human-computer interface, content-based retrieval, and so on. 
Although there have been many works during the past decades \cite{Ryoo2009,Rapitis2013,Aliakbarian2017,Wang2017,Vahdat2011,Yun2012}, it is still a challenging problem, especially when the videos offer unconventional conditions, such as unusual viewpoints and cluttered background.

Most of the previous works focused on mutual action recognition using RGB videos~\cite{Ryoo2009,Rapitis2013,Aliakbarian2017,Wang2017}.
These solutions comprise approaches going from hand-crafted local features to data-driven feature extraction~\cite{Ryoo2011,Zhang2012,Donahue2015,Aliakbarian2017,Shi2018}.
Some methods try to implicitly learn the information from the poses of the persons interacting~\cite{Vahdat2011,Rapitis2013}.
These solutions often operate over the pixel data, making use of manually annotated regions or poselet labels to focus on these specific regions on the frame or to add extra ad-hoc information.
However, they fail to explicitly model the relationships between the interacting body parts of the involved persons, crucial information for interaction recognition.

With the development of more advanced capturing devices that can also capture depth and extract pose information from the people in the scene (e.g., Microsoft Kinect~\cite{Zhang2012a}), a more accurate and valuable description for the human pose can be directly used by new solutions.
Besides, very recently, techniques that can estimate poses from regular RGB videos have improved significantly~\cite{Cao2018}. This allows us to apply reliable pose-based solutions to mutual action recognition in the RGB videos.

The availability of explicit pose representation of the humans in the videos led to a new branch of interaction recognition techniques focusing on this type of data~\cite{Yun2012,Ji2014,Ji2015,Li2016,Wu2018a,Perez2019a}. 
These solutions usually involve hand-crafted features and a classification technique that theoretically incorporates the different body-parts' relationships, and then generate a rigid architecture highly dependent on what is believed to be the prior knowledge over the human skeleton structure.

\begin{figure}[!tp]
	\centering
	\includegraphics[]{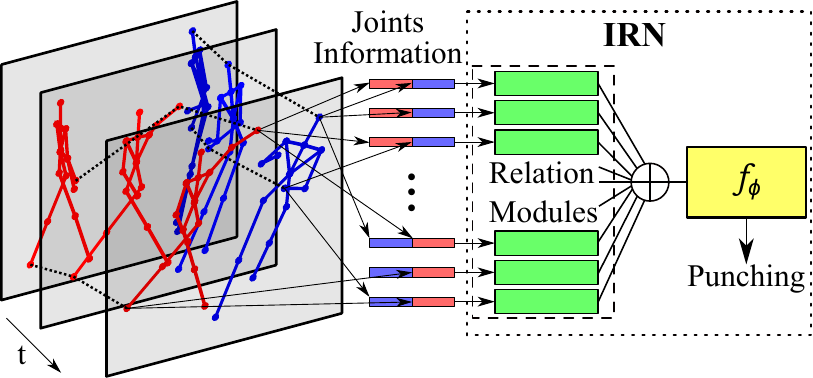}
	\caption{Interaction Relational Network (IRN) overview. Joints of both individuals are separately paired up, and then fed to independent relation modules. The pair-wise inferred relationships are then aggregated into a global description used for, at last, perform human interaction recognition.}
	\label{fig:summarized_pipeline}
\end{figure}

There also exists solutions, using pose information, that target general action recognition~\cite{Du2015,Zhu2016a,Liu2016a,Cai2016,Yang2017,Liu2018,Ke2018a,Ke2019,Liu2019d}, i.e. they include actions performed by a single individual as well. 
Majority of more recent approaches are based on LSTMs and CNNs, usually ad-hoc network architectures, extra regularization or losses, which tries to better model the spatial and temporal dynamics of the poses during the action.
Although these works achieve promising results on general actions, they lack means of leveraging the relationship information between the poses of the two persons when dealing with interactions, under-performing in those cases.

Recurrent models (RNN/LSTM) and convolutional models (CNN/GCN) with carefully designed and complicated architectures have dominated the field of pose-based (skeleton-based) action recognition for several years. 
Instead of using recurrent/convolutional architectures for feature learning, in our paper we propose to view the mutual action recognition task as a body-parts relation inferring and learning problem, and thus design a new architecture to automatically handle this task.
Our contribution is based on the powerful Relational Network (RN)~\cite{Santoro2017} for implicitly reasoning over the joints relationships present at human interactions. 
Although simple, RNs proved to be highly efficient at learning what are the important relationships between pair of objects.
Santoro et al.~\cite{Santoro2017} proposed RNs as a simple architecture to deal with problems that require relational reasoning, obtaining state-of-the-art performance on tasks such as visual and text-based question answering, as well as understanding the dynamic of physical systems that included motion data.
To the best of our knowledge, no one before have developed a solution based on Relational Network with the purpose of interaction recognition, nor with explicit body information.

On this work we propose Interaction Relational Network (IRN), summarized in Fig.~\ref{fig:summarized_pipeline}. 
Since our method is the first attempt inspired by the Relational Networks (RNs) for skeleton-based interaction recognition, our first contribution lies on how to logically rearrange our problem components and characteristics in terms that match the high-level concept behind the original RNs. 
In other words, on how to use the pose information as input, identifying the existing objects (joints) and their properties (coordinates through time), and how to adequately pair these objects, representing the potential relations relevant for interaction recognition. 
In our new architecture, instead of using pre-defined joints dependency structures as previous methods, we enable our network to infer by itself the existing relations among specific relationships types.
We identified two types of relationships, namely, $intra$ relations of body joints from the same person and $inter$ relations of the joints from different persons, thus the complex relations of different body parts in the person-person interaction sequence are well represented. 
The relation modules from the same relationship mapping share the same weights, and will infer from the consciously paired up joints which relations exists between them, and which are important for interaction recognition. 
The relationship mapping, which defines how the joints are paired up, will determine the type of relationships being inferred by that Relation Module ($intra$ or $inter$).
The description generated from all the relation modules will then be pooled and fed to a module, which will learn how to interpret this arrangement of relations, and classify the interaction accordingly.

Given the specificity of our task and data, we propose two distinct types of relations, however the original RNs contains a single model regardless of relationship type.
Thus another contribution in our work is the design of multi-relationship architectures, merging different relation models into a single end-to-end network and leveraging both types of information, therefore leading to a more accurate recognition. We demonstrate the importance of appropriately choosing a fusion architecture, and initializing it with the right models prior to training.
Furthermore, we enrich our solution by equipping our new network with structured operations over the object-pair, based on domain knowledge, to extract valuable joint distance and motion information before reasoning about their relationships. In contrast, the original RNs only contain pair independent operations previously to the relational module.
At last, we incorporate an LSTM to our architecture, in order to enable the IRN reasoning over the interactions relationships over a longer duration sequence as well.

We validate our approach through extensive experiments on four different datasets.
Two of these are traditional datasets for human interaction recognition: SBU~\cite{Yun2012} and UT~\cite{Ryoo2009}. On which we obtain the state-of-the-art performance, close to perfect accuracy.
Seeking to further validate our IRN with more challenging data, we propose to use the large-scale datasets NTU RGB + D~\cite{Shahroudy2016} and NTU RGB + D 120~\cite{Liu2019}.
Although these datasets were initially created for benchmarking general human action recognition, they contain many classes of mutual actions, in fact more than SBU and UT.
Additionally, they have much more samples per class and adverse evaluation protocols.
Even though the NTU RGB + D datasets are more difficult, the IRN still outperforms the previous works.

Our contributions can be summarized in the following manner:
\begin{itemize}
	\item A novel approach to Human Interaction Recognition, using the different body parts from the pose information as independent objects, and modeling their relationship pairwise. 
	\item We design effective ways to fuse different types of relationships, leveraging their complementary for an improved performance.
    \item We extend the relational network formulation with structured pair-wise mechanisms, that allows it to self-augment its input with relevant extra information from the input pair.
	\item A new and efficient paradigm for interaction recognition, based on relational networks, as an alternative to the CNN/RNN/GCN architectures that currently dominate the field.
\end{itemize}
These contributions constitute the novel architecture Interaction Relational Network, which is simple yet effective. 
We validate on four important datasets that contains human interactions under different conditions, achieving state-of-the-art performance in three of them and competitive results in the fourth, demonstrating the strength of our technique.

This paper is an extension of our preliminary conference work \cite{Perez2019a}.
For the new work in this extension a novel formulation for the relational network with a pair-wise structured input is provided, allowing our IRN to automatically extract distance and motion features.
In this new work, we also propose an improved relationships fusion architecture, which takes leverage from higher-level inferred relations.
Another important new addition in this paper is the coupling of the LSTM to our architecture, enabling our method to perform temporal relational reasoning, and to reason over the entire interaction sequence.
To further assess our proposed method, we perform experiments in two additional datasets with challenging characteristics: UT-Interaction, on which the poses had to be estimated; and NTU RGB+D 120, which contains many classes.
Finally, in this new work we perform a more thoroughly qualitative analysis, with the aid of confusion matrices and a bar chart for performance per interaction class.

\section{Related Work} %
\label{sec:related_work}

\subsection{Action Recognition}

The problem of human interaction recognition is in fact a sub-field from the more general task of action recognition, which also takes into consideration actions performed by a single person.
Previous work with methods using solely RGB videos usually rely on spatio-temporal local features description~\cite{Wang2013} or CNN architectures for processing appearance and motion separately~\cite{Simonyan2014,Wu2016,Shi2017a}.
Here we will focus on the previous work that also uses explicit pose information, and have as well evaluated their methods on the SBU interaction dataset.

Most of the latest works using pose information have moved to solutions involving Deep Learning, specially LSTMs~\cite{Du2015,Zhu2016a,Liu2016a,Liu2018}.
Du et al.~\cite{Du2015} designed a hierarchical structure using Bidirectional Recurrent Neural Networks (BRNNs), which initially process the joints separately by body part, then fusing these parts step-by-step at subsequent layers, until the whole body is covered.
Zhu et al.~\cite{Zhu2016a} introduced a regularization in their LSTM to learn the co-occurrence features from the joints -- i.e. learn the discriminative joint connections related to human parts.
Liu et al.~\cite{Liu2016a} proposed a spatio-temporal LSTM architecture, where the network would unroll not only in the temporal domain, with sequential frames, but also in the spatial, with a pre-defined traversal order of the joints following the human skeleton structure.

Since these works focus on general actions, which mostly are performed by a single person, they do not include in their solutions any strategy specifically to model the relationship between the interacting persons when it is a mutual action.
Therefore they disregard the additional information coming from the interaction itself, which our experiments demonstrate to be invaluable when dealing with two-person action recognition.

\subsection{Human Interaction Recognition}

Most of early work on human interaction recognition is based on hand-crafted local features, following a bag-of-words paradigm, and often using SVM for the final classification~\cite{Ryoo2009,Vahdat2011,Ryoo2011,Zhang2012,Rapitis2013}.
More recent and successful approaches rely on Convolutional Neural Networks for feature extraction, usually combining it with Recurrent Neural Networks (e.g., Long-short Term Memory) for more complex temporal reasoning over the interactions~\cite{Donahue2015,Aliakbarian2017,Shi2018,Ke2018}.

Some of these approaches allegedly focus on pose~\cite{Vahdat2011,Rapitis2013}, or try to embed human structural knowledge on their solution~\cite{Ke2018}.
Vahdat et al.~\cite{Vahdat2011} 
requires pre-computed person trajectories for their solution, initially they employ a pedestrian detector to obtain bounding boxes around the individuals on the first frame, then subsequently apply a tracker to find the trajectory in the later frames.
These bounding boxes are used to retrieve the input information for their solution: spatial location, time of the pose and the cropped image within this region.
On the work of Rapitis et al.~\cite{Rapitis2013}, the authors have labeled pre-defined poselets on some of the frames. They create templates from these poselets by extracting local-features on the cropped image around these annotations. 
Based on these poselets features, their max-margin framework would then select key-frames to model an action.

Another branch of works do use explicit pose information with the purpose of interaction recognition~\cite{Yun2012,Ji2014,Ji2015,Wu2018a}. 
For example Yun et al.~\cite{Yun2012} proposed features coming from different geometric relations between the joints intra-person and inter-person, as well as intra-frames and inter-frames, then using these features as input to an SVM for classification.
Ji et al.~\cite{Ji2014} grouped joints that belonged to the same body part to create poselets, for describing the interaction of these body parts from both individuals, subsequently generating a dictionary representation from these poselets, which will then be fed to an SVM.

Differently from the above mentioned approaches, our proposed architecture does not need the prior knowledge about the skeleton structure.
We do not use the edge information in our architecture and the input is invariant to the joint order, therefore it learns the existing and important relationships by itself.

\subsection{Relational Network}

Relational networks have been recently introduced by Santoro et al.~\cite{Santoro2017}, targeting to improve relational reasoning over a different range of tasks. 
It consists of initially reducing the problem input to individual objects, then using a simple architecture that models how these objects relate to each other in a pair-wise form and subsequently uses the general view from all the inferred relationships to solve the problem at hand.
The authors demonstrated the versatility of these networks by applying their method on three tasks: visual question answering (QA), text-based question answering and complex reasoning about dynamic physical systems. 
Which covered not only distinct purposes (question-answering and physical modeling) but also different types of input data (visual, textual and spatial state-based).

\begin{figure*}[!t] 
	\centering
	\includegraphics[width=\textwidth]{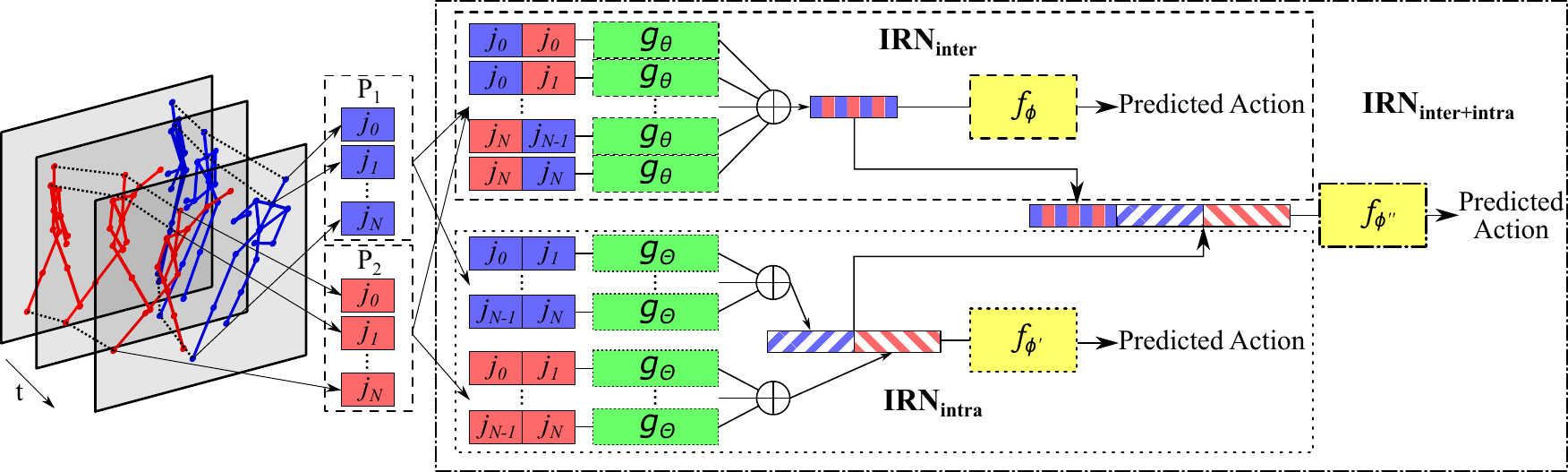}
	\caption{Illustration of the IRN architecture for human interaction recognition. First, we extract the information across frames from each joint separately ($j_n$). Then we use the set of joints from both persons ($P_p$) as input to our different architectures, $IRN_{inter}$ and $IRN_{intra}$. Each architecture models different relationships between the joints, and can be used independently to predict the action. Furthermore, the models can be fused as $IRN_{inter+intra}$, this way using the knowledge from both types of relationships 
	for a more accurate prediction. 
}
	\label{fig:detailed_pipeline}
\end{figure*}

Expanding RNs to Video QA, Chowdhury et al.~\cite{Chowdhury2018} proposed Hierarchical Relational Attention. 
On which the authors connected attention modules over VGG and C3D features to hierarchical relational modules, nested by input question query token.
Park and Kwak~\cite{Park2018} applied RNs for 3D Pose Estimation. In their work they used as input estimated 2D joints coordinates, that are grouped by body part and fed in a specific order to their proposed RN to generate the estimated 3D position.
Ibrahim et al.~\cite{Ibrahim2018} also employed Hierarchical RNs, but this time for group activity recognition. 
They start from CNN features from each individual, then input these features to hierarchically stacked multiple layers of relational modules, each one with their own parameters and relationships graph, which dictates the pairwise relationships to be made.

As far as we know, none of the previous work have designed relational networks that use pose information for the purpose of Human Interaction Recognition, with our work being the first one to extend RNs for this domain and application.

\section{Relational Network Overview} %
\label{sec:related_concepts}

The Relational Network proposed by Santoro et al.~\cite{Santoro2017} can be simplified through the following equation:

\begin{equation}
	RN(O)=f_\phi\left(\sum\limits_{i,k}{g_\theta\left(o_i,o_k\right)}\right)\label{eq:rn}
\end{equation}

Where $O$ is a set of objects $\{o_1,o_2,...,o_N\}$, 
on which any $i^{th}$ object is represented by an arbitrary $\mathbb{R}^m$ vector containing the properties of that object. The function $g$, with learnable parameters $\theta$, is responsible by modeling the relationships for each pair of input objects independently, therefore being also referred to as Relational Module.
Meanwhile function $f$, with trainable parameters $\phi$, is in charge for reasoning from the merger of the relationships inferred by $g_\theta$.
It is a versatile formulation which, as long as the objects are fed in pairs to the relational module, can have as input different types of data describing the objects, such as CNN/LSTM features or even physical states.

However, the original RN does not cover the case when well-distinguished relationships can be defined due to the nature of the objects and the chosen pairing up rule, leading to distinct specialized models for each type. Missing therefore good ways to train these models and properly fusing them afterward. 
Also, in its base formulation the Relational Network only contains an unstructured function for reasoning over the input pair, represented by $g_\theta$, lacking a pair-wise structured function that can guide the general relational module with important domain-knowledge information.
In the process of tailoring the RN paradigm to our domain, we design important novel extensions to address these issues on our proposed Interaction Relational Network, detailed next.

\section{Interaction Relational Network}
\label{sec:prop_method}

We designed an architecture, namely Interaction Relational Network, tailored for human interaction recognition using pose information, inspired by the relational network paradigm of reasoning over pair-wise relationships.
Fig.~\ref{fig:detailed_pipeline} contains a visual representation of the proposed method.
First we define how to logically re-arrange the skeleton information so it can be used as input to an RN based architecture, then we formulate important relationships mapping as to comprise all relevant relations pertaining to our problem.
In order to fully capture the range of relationships types present in the human interaction problem, it was necessary to extend the RN proposal to handle multiple relations models and to fuse them properly.
Also, we further extend the architecture with domain-knowledge pair-wise operations over the input at each relation module, allowing the network to explicitly extract information known to be valuable when dealing with pose data: distance and motion.
Finally, we couple our architecture with an LSTM, allowing our method to reason over the interactions during the whole video sequence.

\subsection{Skeleton Joints as Independent Objects}

The power of the relational networks lies on breaking down the problem into separate interacting objects, and learning specific parameters for describing how each object relate to each other, before merging this information for the classification of the whole.
In the case of pose information and action recognition, we can define each joint $j_i$ as an object, using as their low-level feature its coordinates along the frames, together with integers to identify to which joint and body part does it belong: $j_i=(x_1,y_1,x_2,y_2,...,x_T,y_T,i,b)$, where $x_t$ and $y_t$ are the 2D coordinates of the joint $i$ belonging to the body part $b$ at the frame $t$, and $T$ is the desired sampling of frames to be used.
As previous work have done \cite{Du2015,Shahroudy2016,Liu2018}, we considered five body parts: torso, left hand, right hand, left leg, and right leg.
Each person $p$ will therefore have a set of joints for each video, which can be defined as: $P_p=\{j^p_1,j^p_2,...,j^p_N\}$, where $N$ is the total number of joints provided by the pose data.

\subsection{Relationships Architectures}

The simplest approach would be to mix and match all the joints, learning a single model with all pairs combinations.
However, besides being computationally inefficient, this naive strategy disregards the fact that the joints might come from other individuals, hence the relationship which each other carries different meaning than when the joints are from the same individual.
To allow our method to distinguish between the relevant relationships, and therefore better represent human interactions, we propose exclusive relationships mappings, which will be later merged into a single architecture.

\textbf{Inter-person Relationships.}
Since we are dealing with interaction recognition, it is desired to map the relation between the joints inter-person.
Therefore, all joints from one individual will be paired with all joints from the other individual, bi-directionally, so that it can be order-invariant (in the case of active/passive actions such as kicking).
For that purpose, inspired by the formulation (\ref{eq:rn}), we derive the following equation:

\begin{equation}
IRN_{inter}(P_1,P_2)=f_\phi\left(\sum\limits_{i,k}{g_\theta\left(j^1_i,j^2_k\right)} \oplus \sum\limits_{i,k}{g_\theta\left(j^2_i,j^1_k\right)}\right)
\end{equation}

Where $f_\phi$ and $g_\theta$ can be Multi-Layer Perceptrons (MLPs), with learnable parameters $\phi$ and $\theta$ respectively. In theory $\sum$ and $\oplus$ can be any pooling operation, such as sum, max, average or concatenate, but from our experiments, we decided to use average because it gives the best results.

\textbf{Intra-person Relationships.}
Since the intra-person relationship of the joints can be highly informative as well, we also propose another architecture, where the joints from each person will be paired with the other joints from the same person. 
For this case there is no need to pair bi-directionally, since the paired joints are from the same person, what would only add unnecessary redundancy to our model -- 
in fact our preliminary experiments demonstrated that it can lead to overfitting in some cases.
The pooled output from each individual is concatenated ($\frown$) before going through 
function $f$ with trainable parameters $\phi'$.

\begin{multline}
IRN_{intra}(P_1,P_2)= \\ f_{\phi'}\left(\sum\limits_{i=1}^{N}\sum_{k=i+1}^{N}{g_\Theta\left(j^1_i,j^1_k\right)} \frown \sum\limits_{i=1}^{N}\sum_{k=i+1}^{N}{g_\Theta\left(j^2_i,j^2_k\right)}\right)
\end{multline}

\textbf{Fusing Relationships.}
Conclusively, we propose an architecture that fuses both types of relationships under the same 
function $f$ (parameters $\phi''$), 
by concatenating the pooled information from each function $g$, each with its own parameters $\theta$ and $\Theta$:

\begin{multline}
IRN_{inter+intra}(P_1,P_2)= \\ 
f_{\phi''}\left(\sum\limits_{i,k}{g_\theta\left(j^1_i,j^2_k\right)} \oplus \sum\limits_{i,k}{g_\theta\left(j^2_i,j^1_k\right)} \frown \right. \\
\left. \sum\limits_{i=1}^{N}\sum_{k=i+1}^{N}{g_\Theta\left(j^1_i,j^1_k\right)} \frown \sum\limits_{i=1}^{N}\sum_{k=i+1}^{N}{g_\Theta\left(j^2_i,j^2_k\right)}\right)
\end{multline}

\begin{figure}[t]
	\centering
	\includegraphics[]{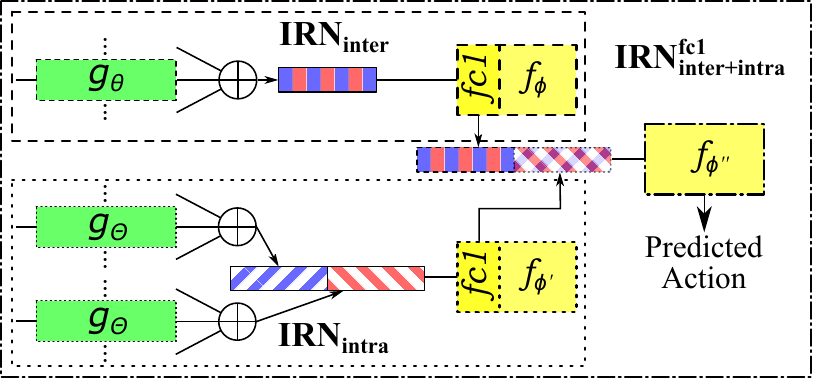}
	\caption{
	Simplified view of $IRN^{fc1}_{inter+intra}$, an alternative way from fusing both types of relationships. On this case, the fusion occurs after the first fully-connected layer ($fc1$) from $f_\phi$ and $f_{\phi'}$, which provides a higher level description for the fusion $f_{\phi''}$.
	}
	\label{fig:fc1_fusion}
\end{figure}

Alternatively, we also designed another fusion architecture: $IRN^{fc1}_{inter+intra}$ (Fig.~\ref{fig:fc1_fusion}).
At this design we concatenate the output from $f$ first fully-connected layer ($fc1$), of $inter$ and $intra$ models ($\phi$ and $\phi'$ parameters respectively), before feeding into the fusion $f_{\phi''}$.
The motivation behind this design is to allow us to transfer knowledge not only from the relational modules $g_\theta$ and $g_\Theta$, but also from their exclusive $f$ modules.
The modules $f_{\phi}$ and $f_{\phi'}$ should be specialized on the pooled descriptor for its respective type of relationships, hence contributing with a higher level description for the fusion.

\subsection{Pair-wise Structured Input}

One of the advantages of RNs highlighted by Santoro et al.\cite{Santoro2017} is the flexibility it has to reason over relatively unstructured inputs, for example embeddings from CNNs and LSTMs, as well as it can handle raw data such as state descriptors and, in our case, joints coordinates.
Nonetheless, when dealing with structured data it is advantageous to design the architecture with mechanisms to leverage the intrinsic information contained in the data structure.
Although this is the case with CNNs and LSTMs used for extracting the embeddings used as input to the RN, they are applied considering only the structure per-object independently.
In the case of joints coordinates, the input pair itself is organized in a manner that there is a well-defined structure among them, i.e. coordinates at the same location in the array represents the spatial location of the objects at the same particular time in the video sequence.
Therefore it can also be designed architectures to take leverage of the pair-wise structure of the input.

In order to enable the RN to allow this desired feature, but also keeping its flexibility, we propose the following extension to the base formulation (\ref{eq:rn}), on which function $h$ can be carefully designed to model important domain-knowledge pair-wise information:

\begin{equation}
	RN(O)=f_\phi\left(\sum\limits_{i,k}{g_\theta\left(o_i,o_k,h\left(o_i,o_k\right)\right)}\right)\label{eq:rn_h}
\end{equation}

Inspired by previous work that demonstrated the usefulness of explicit distance and motion features for pose-based action recognition~\cite{Yun2012,Wu2018a}, we decided to augment our relational network with modules that can automatically extract this information from each pair of joints. 
In other words, our $h$ function, to be also used as extra input by $g_\theta$, will explicitly compute vectors of distances ($D(j_i,j_k)$) and motions ($M(j_i,j_k)$) from the input object-pair:

\begin{equation}
h\left(j_i,j_k\right)=\left(D(j_i,j_k) \frown M(j_i,j_k)\right)
\end{equation}

Considering $c^i_t$ the coordinates of joint $i$ at the frame $t$, the vector of distances is basically the euclidean distance between the input joints at each frame.

\begin{equation}
D(j_i,j_k)=(\left\|c^i_1-c^k_1\right\|,\left\|c^i_2-c^k_2\right\|,...,\left\|c^i_T-c^k_T\right\|)
\end{equation}

And the motion vector is defined as the distance between the joints, but at sequential frames:

\begin{equation}
M(j_i,j_k)=(\left\|c^i_1-c^k_2\right\|,\left\|c^i_2-c^k_3\right\|,...,\left\|c^i_{T-1}-c^k_T\right\|)
\end{equation}

So we can write the improved $g_\theta$, used by $IRN^{'}$, as: 

\begin{equation}
g^{'}_\theta\left(j_i,j_k\right)=g_\theta\left(j_i,j_k,h(j_i,j_k)\right)
\end{equation}

\subsection{Sequential Relational Reasoning}

Until this point, our proposed IRN could only reason on the interactions over a short period of time.
Defined by the number of frames ($T$) sampled when assembling the joints information ($j_i$) for input to our network.
In other words, the hyper-parameter $T$ would be analogous to a temporal receptive field of our IRN.
Preliminary experiments showed that using a too large value of $T$ is detrimental to performance.
It was more effective to skip some of the frames by some step size, as in dilated convolutions~\cite{Yu2016}, in order to increase the temporal range of the input.
However, this workaround still does not allow the IRN to use the information available in the whole video sequence, and moreover, it has no means to model the intrinsic order of movements contained in an interaction.

Seeking to enable our method to use information from all frames and reason over the evolution of the interactions over time, we incorporate an LSTM to our architecture. %
The LSTM module is at the end of our architecture, after $f_{\phi}$ and before predicting the action.
This architecture, hereinafter referenced as LSTM-$IRN$, can then leverage the information from the whole interaction sequence before classifying the video.
Resulting in an architecture also suitable to model over time the relationships present in human interactions.

\section{Experiments}
\label{sec:experiments}

\subsection{Datasets}

\textbf{SBU}~\cite{Yun2012} is a dataset for two-person interaction recognition, created using Kinect, providing reliable RGBD data for each video. 
It consists of eight interactions (approaching, departing, pushing, kicking, punching, exchanging objects, hugging, and shaking hands), seven different participants (pairing up to 21 different permutations), on a total of 282 short videos (around 2-3 seconds each). 
Recording was done in a single laboratory environment and with a frame rate of 15 frames per second (FPS).
Pose information is provided by means of 3D coordinates over 15 joints per person, at each frame.
Coordinates are not entirely accurate, containing noise and incorrect tracking at some cases.
We followed the 5-fold cross validation protocol defined by the authors, reporting the average accuracy.

\textbf{UT-Interaction}~\cite{Ryoo2009} is also a dataset that focus on inter-person interaction, but differently from SBU, it only contains RGB information -- i.e. no explicit pose information is provided. 
To overcome this issue we used OpenPose~\cite{Cao2018} to extract this type of information, estimating the joints coordinates for the actors in the video.
This dataset contains six different classes of actions (shake-hands, point, hug, push, kick, and punch). It was recorded under two different set of conditions, therefore it is subdivided into UT-1 and UT-2, comprising a total of 120 videos, half at each set. 
Set 1 was recorded on a parking lot, with static background and little camera jitter, meanwhile, Set 2 was taken on a lawn, with partial occlusion from tree branches and with interference from the wind (camera jitter and moving leaves).
Videos have about 2-6 seconds each, and a frame rate of 30 FPS.
Authors protocol for evaluation consists of a 10-fold cross validation per set.

\textbf{NTU RGB + D}~\cite{Shahroudy2016} is a dataset with a wide range of general actions.
It is not a dataset exclusively for interaction recognition, however it contains 11 classes of mutual actions (punch/slap, pat on the back, giving something, walking towards, kicking, point finger, touch pocket, walking apart, pushing, hugging, handshaking), more than SBU and UT-Interaction. 
Moreover, this subset with interaction-only classes, contains a total of 10,347 videos, which were collected with the more precise Kinect (V2) and under more challenging conditions, with 40 different subjects and large variation in viewpoints, by using three cameras recording at the same time.
The length of the videos range from 1 to 7 seconds, with a frame rate of 30 FPS.
The dataset contains the 3D coordinates from 25 joints per person for all frames.
For evaluation, the authors proposed two protocols: Cross Subject (CS), on which 20 pre-defined actors are used for training, and the remaining for testing; and Cross View (CV), where two-cameras are reserved for training, 	
Although this is not a dataset specifically for Human Interaction Recognition, we believe that experimenting over this dataset mutual-only classes can be highly valuable at validating our methods because of its characteristics: large scale and more challenging conditions.

\textbf{NTU RGB + D 120}~\cite{Liu2019} is an extension from the dataset above, and it is the largest available dataset for human action recognition with RGBD data.
It contains 60 additional classes, of which 15 are mutual-action classes (hit with object, wield knife, knock over, grab stuff, shoot with gun, step on foot, high-five, cheers and drink, carry object, take a photo, follow, whisper, exchange things, support somebody, rock-paper-scissors). 
In total there are 24,794 videos from 26 interaction-only classes, captured similarly as the previous version: using Kinect (V2), with many different subjects and a diverse range of viewpoints.
The evaluation protocol is slightly different: Cross-Subject, which is the same as before, with a subset of actors being used exclusively for training and another for testing; and Cross-Setup, where instead of splitting the data according to the camera, pre-defined setups are selected for comprising the training and testing splits.

\begin{figure*}[!ht]
	\centering
	\includegraphics[trim={6 2 10 0},clip]{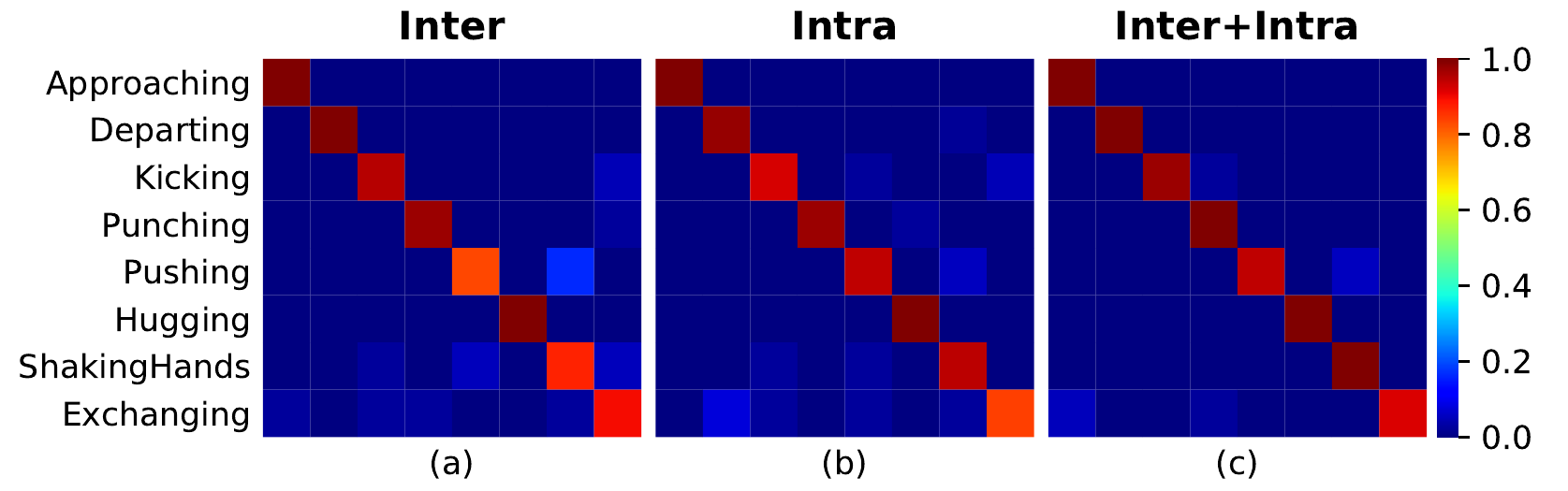}
	\caption{Confusion matrices for SBU dataset with methods: (a) LSTM-$IRN^{'}_{inter}$, (b) LSTM-$IRN^{'}_{intra}$ and (c) LSTM-$IRN^{'fc1}_{inter+intra}$.
	}
	\label{fig:confusion_matrix_sbu}
\end{figure*}

\subsection{Implementation Details}

\textbf{MLPs Configuration.}\quad Hyperparameters detailed here were tuned during preliminary experiments.
The IRN is implemented as an MLP, where $g_\theta$ consists of four fully-connected layers, the first three with 1000 units and the last with 500, and
$f_\phi$ contains three fully-connected layers with 500, 250 and 250 units respectively, with dropout rate of 0.25.
The LSTM module contains 256 units and also a dropout of 0.25.
A softmax layer is placed at the end of the architecture to generate the interactions prediction scores.
Training was performed with Adam optimizer, learning rate value of 1e-4 and weight initialization following a truncated normal distribution with zero mean and 0.045 standard deviation for SBU and UT, and 0.09 standard deviation for NTU RGB+D and NTU~RGB~+~D~120.

\textbf{Training Procedure.}\quad To improve generalization during training, we randomly swap the input order between the persons' joints set ($P_1\rightleftharpoons{P_2}$). This was significantly beneficial for the $IRN_{intra}$ architecture to avoid bias on the order of the concatenated feature generated after $g_\Theta$.
$IRN_{inter+intra}$ parameters $\theta$ and $\Theta$ are fine-tuned from the weights obtained previously by training $IRN_{inter}$ and $IRN_{intra}$ separately, meanwhile $\phi$ is randomly initialized.
For $IRN^{fc1}_{inter+intra}$,  in addition to $\theta$ and $\Theta$, 
we also use the $fc1$ layer and weights from previously trained $IRN_{inter}$ and $IRN_{intra}$, for initializing the model before training.

\textbf{Pose Estimation.}\quad OpenPose~\cite{Cao2018} is a readily available toolbox for extracting pose information from RGB videos. We ran it with its default options, which were not the most accurate, but it was faster and sufficiently precise for our experiments.
However, because the output is frame-based, we had to apply some post-processing steps to guarantee the consistency of the bodies from Person 1 and 2 throughout the video.
It consisted mainly on correctly assigning the pose at each frame to its respective body, based on the distance of the joints between the current frame and previous frames, assuming the overall distance should not be too different between consecutive frames.
This is important when there are more than two poses in the frame, due to noise or because of passersby, or when the order of the actors' pose change.

\textbf{Joints and Frames Subsampling.}\quad 
Although the poses for UT (estimated) and NTU RGB+D (provided) contains 25 joints, we sampled only 15 of them, analogous to what is provided by SBU data. 
As for the parameter T, regarding the temporal receptive field, we use different values for each dataset.
For SBU, since the videos are shorter, we use 8 consecutive frames as a sampling for our input feature. %
Since UT, NTU RGB+D and NTU RGB+D 120 have longer videos, we use 32 frames. 
For the experiments without an LSTM we have chosen to sample the central frames, because most likely they contain the more relevant parts of the interaction.
For LSTM-IRN we build our input sequence with overlapping frames at each timestep, by sampling T frames of the video with a step of $T/2$ frames, i.e. 8 frames input for every 4 frames step for SBU and 32 frames every 16 for the other datasets.

\subsection{Ablation experiments}

First, to better evaluate the impact of each part from our proposed methodology, we separately show our results on the SBU dataset, reported in Table~\ref{tab:sbu_results}.

\begin{table}[ht]
	\begin{center}
	\setlength{\tabcolsep}{2em}
	\renewcommand{\arraystretch}{1.4} %
	\caption{Ablation experiments on SBU dataset.}
	\begin{tabular}{lc}
		\toprule
		\textbf{Experiment} 						& \textbf{Acc} \\ \midrule
		$IRN_{inter}$ (Baseline)					& 88.7\%			\\ 
		$IRN^{'}_{inter}$ (Self-Augmented Input)	& 93.6\%			\\ 
		LSTM-$IRN^{'}_{inter}$						& 94.6\%			\\ \midrule
		
		$IRN_{intra}$ (Baseline)					& 95.4\%			\\ 
		$IRN^{'}_{intra}$ (Self-Augmented Input)	& 95.8\%			\\ 
		LSTM-$IRN^{'}_{intra}$						& 95.2\%			\\ \midrule
		
		Naive-$IRN^{'}_{inter+intra}$		        & 92.0\%	\\ 
		Averaging scores                            & 94.3\%	\\ 
		Random-$IRN^{'}_{inter+intra}$	            & 92.5\%	\\ 
		$IRN^{'}_{inter+intra}$                     & 96.1\%	\\ 
		$IRN^{'fc1}_{inter+intra}$		        	& 96.7\%	\\ 
		$IRN^{'fc2}_{inter+intra}$		            & 94.9\%	\\ 
		$IRN^{'fc3}_{inter+intra}$		            & 95.3\%	\\ 
		LSTM-$IRN^{'fc1}_{inter+intra}$	            & 98.2\%	\\ \bottomrule
	\end{tabular}
	\label{tab:sbu_results}
	\end{center}
\end{table}

Our baseline, using only the central frames coordinates and joint indexing information, already obtains 88.7\% of accuracy with $IRN_{inter}$ and 95.4\% of accuracy with $IRN_{intra}$ architecture. These results indicates that our approach is able to successfully map the different types of relationships present in the problem of interaction recognition.
Incorporating our IRN with mechanisms to take leverage of the pair-wise structure of the input, what allows the network to extract explicit distance and motion information, proved to be highly valuable to $IRN_{inter}$ and it was also helpful to $IRN_{intra}$, increasing the performance on both cases.

The Naive-$IRN^{'}_{inter+intra}$ experiment consists on simply merging all the relationships mapping ($inter$ and $intra$) into a single model relational network. 
This means only one set of parameters for $g$ and pooling the outputs indiscriminately into the same global descriptor before $f$. 
The Naive-$IRN^{'}_{inter+intra}$ approach underperforms, obtaining only 92.0\% of accuracy, lower than the specialized $IRN^{'}$ models.
Averaging the scores from these specialized models is better than naively merging the relationships, but it is still worse than using $IRN^{'}_{intra}$ alone.

We were able to truly take leverage of the complementary between the relationships models only through our proposed fused architecture $IRN^{'}_{inter+intra}$. This method maintains the specialized models not only on its design, but also by initially training them separately -- random initialization of weights (Random-$IRN^{'}_{inter+intra}$) also underperforms.
Furthermore, our $IRN^{'fc1}_{inter+intra}$ approach seems to further correlate and benefit from the complementary of the relationships, obtaining a more significant improvement.
However, our attempts to fuse at even higher-levels of the MLPs (fc2 and fc3 layers) was not beneficial, due to higher overfit. Thus fc1 is the optimal layer for fusing the relationships.

Ultimately, we extend the relational reasoning to all available frames and enable reasoning over the interaction sequence by coupling our IRN with the LSTM module. 
Although leading to higher over-fitting for the $intra$ relationship type on this small-scale dataset, which obtains a slightly lower performance than the baseline, the LSTM addition is advantageous for the $inter$ variation and, more importantly, when fusing the relationships.
With LSTM-$IRN^{'fc1}_{inter+intra}$, our method achieves the best performance in this dataset: 98.2\%.

Fig. \ref{fig:confusion_matrix_sbu} contains the confusion matrices for all LSTM-$IRN$ architectures. 
It is interesting to notice how the two relationships models have confusion on different interactions, for example $Inter$ have some confusion between \textit{Punching} and \textit{Exchanging} while $Intra$ does not, and how this confusion is greatly reduced at $Inter+Intra$.
More importantly, interaction classes getting confused by both models, such as \textit{Pushing} and \textit{ShakingHands}, are almost entirely distinguished with $Inter+Intra$. 
This qualitative analysis is a good indication on the capability of our proposed architecture to keep the strength of both types of relationships models, and also to leverage their complementary for distinguishing between even harder cases.

\subsection{State-of-the-art Comparisons}

\begin{table}[ht]
	\begin{center}
	\setlength{\tabcolsep}{2em}
	\renewcommand{\arraystretch}{1.4} %
	\caption{Comparison of our results with previous work on SBU.}
	\begin{tabular}{lc}
		\toprule
		\textbf{Method}                                   & \textbf{Acc} \\ \midrule
		Yun et al.~\cite{Yun2012}                         & 80.3\% \\ 
		Ji et al.~\cite{Ji2014}                           & 86.9\% \\ 
		HBRNN~\cite{Du2015} (reported by \cite{Zhu2016a}) & 80.4\% \\ 
		CHARM ~\cite{Li2015}                              & 83.9\% \\ 
		CFDM~\cite{Ji2015}                                & 89.4\% \\ 
		Co-occurrence LSTM~\cite{Zhu2016a}                & 90.4\% \\ 
		Deep LSTM (reported by ~\cite{Zhu2016a})          & 86.0\% \\ 
		ST-LSTM~\cite{Liu2016a}                           & 93.3\% \\ 
		VA-LSTM~\cite{Zhang2017}                          & 97.2\% \\ 
		Wu et al.~\cite{Wu2018a}                          & 91.0\% \\ 
		Two-stream GCA-LSTM~\cite{Liu2018}                & 94.9\% \\
		\midrule
		LSTM-$IRN^{'}_{inter}$														& 94.6\% \\ 
		LSTM-$IRN^{'}_{intra}$														& 95.2\% \\ 
		LSTM-$IRN^{'fc1}_{inter+intra}$                   & 98.2\% \\
		\bottomrule
	\end{tabular}
	\label{tab:sbu_sota}
	\end{center}
\end{table}

\textbf{SBU.}\quad We report our best results for SBU dataset in Table~\ref{tab:sbu_sota}, alongside previous work results. 
Our specialized relationship architectures, LSTM-$IRN^{'}_{inter}$ and LSTM-$IRN^{'}_{intra}$, can already outperform or be comparable to almost all of previous work.
After fusion through architecture LSTM-$IRN^{'fc1}_{inter+intra}$ we obtain state-of-the-art performance, with an advantage over 1 percentage point over the second best: 97.2\% with VA-LSTM~\cite{Zhang2017}.

\begin{table}[ht]
	\begin{center}
	\setlength{\tabcolsep}{1em}
	\renewcommand{\arraystretch}{1.4} %
	\caption{Results from previous approaches and our proposed methods on the UT dataset per set.}
	\begin{tabular}{lcc} 		\toprule
		& \multicolumn{2}{c}{\textbf{Acc}} \\
		\textbf{Method} & \textbf{UT-1} & \textbf{UT-2} \\ \midrule
		Ryoo et al.~\cite{Ryoo2009}								& 70.8\% & -		 \\
		Raptis et al.~\cite{Rapitis2013}					& 93.3\% & -		 \\
		Donahue et al.~\cite{Donahue2015}					& 85.0\% & -			\\
		Kong and Fu~\cite{Kong2016}								& 93.3\% & 91.7\% \\
		Wang and Ji~\cite{Wang2017}								& 95.0\% & -			\\
		Aliakbarian et al.~\cite{Aliakbarian2017}	& 90.0\% & -			\\
		Ke et al.~\cite{Ke2018}										& 93.3\% & 91.7\% \\ 
		Shi et al.~\cite{Shi2018}									& 97.0\% & -			\\ 
		Shu et al.~\cite{Shu2018}									& 98.3\% & -			\\
		\midrule
		LSTM-$IRN^{'}_{inter}$          					& 93.3\% & 96.7\% \\ 
		LSTM-$IRN^{'}_{intra}$          					& 96.7\% & 91.7\% \\ 
		LSTM-$IRN^{'fc1}_{inter+intra}$ 					& 98.3\% & 96.7\% \\ 
		\bottomrule
	\end{tabular}
	\label{tab:ut_sota}
	\end{center}
\end{table}

\textbf{UT-Interaction.}\quad Table~\ref{tab:ut_sota} shows our results on the UT dataset, together with the results from previous works. 
The LSTM-$IRN^{'fc1}_{inter+intra}$ obtains a performance equivalent to the current state-of-the-art on the set UT-1, an almost saturated accuracy of 98.3\%.
On UT-2 set, our architectures outperform the previous methods, and set the new state-of-the-art to 96.7\%, which is also very high. For UT-2 there was no improvement from using LSTM-$IRN^{'fc1}_{inter+intra}$, when in comparison to the specialized LSTM-$IRN^{'}_{inter}$ model, it might be the case that for this dataset the interactions are not that complex, therefore not benefiting so much from more advanced relationships modeling.
To the best of our knowledge, our work is the first using explicit pose information at UT-Interaction dataset, with the previous works being based on RGB information only.
Therefore, our experiments demonstrate that it is also possible to perform high performance human interaction recognition in RGB videos using estimated poses.

\begin{table}[ht]
	\begin{center}
	\setlength{\tabcolsep}{1em}
	\renewcommand{\arraystretch}{1.4} %
	\caption{Results from our proposed methods on the subset of NTU RGB+D dataset, containing only the mutual action classes.}
	\begin{tabular}{lcc} 		\toprule
		& \multicolumn{2}{c}{\textbf{Acc Mutual Actions}} \\
		\textbf{Method} & \textbf{Cross-Subject} & \textbf{Cross-View} \\
		\midrule
		ST-LSTM~\cite{Liu2016a}    				& 83.0\% & 87.3\% \\ 
		GCA-LSTM~\cite{Liu2017c}    			& 85.9\% & 89.0\% \\ 
		Two-stream GCA-LSTM~\cite{Liu2018}      & 87.2\% & 89.9\% \\
		FSNET~\cite{Liu2019c}					& 74.0\% & 80.5\% \\
		ST-GCN~\cite{Yan2018a}		            & 83.3\% & 87.1\% \\
		AS-GCN~\cite{Li2019}			        & 89.3\% & 93.0\% \\
		\midrule
		LSTM-$IRN^{'}_{inter}$         				& 89.5\% & 92.8\% \\
		LSTM-$IRN^{'}_{intra}$         				& 87.3\% & 91.7\% \\
		LSTM-$IRN^{'fc1}_{inter+intra}$				& 90.5\% & 93.5\% \\ 
		\bottomrule
	\end{tabular}
	\label{tab:ntu_inter}
	\end{center}
\end{table}

\textbf{NTU RGB+D.}\quad 
Our experiments results over the interaction classes of this dataset are present in Table~\ref{tab:ntu_inter}.
Both of the specialized architectures are able to match or outperform the majority of the compared methods on the two protocols, including the graph-based method ST-GCN~\cite{Yan2018a}. 
LSTM-$IRN^{'}_{inter}$ by itself obtains a performance very similar to another recent graph-based method AS-GCN~\cite{Li2019}, demonstrating the importance of considering the inter-person joints relationships when dealing with mutual actions.
Our LSTM-$IRN^{'fc1}_{inter+intra}$ architecture can obtain an even higher performance, going beyond 90\% accuracy not only on the Cross-View protocol, but also on the more challenging Cross-Subject, ultimately outmatching all the compared methods in this dataset.

\begin{table}[ht]
	\begin{center}
	\setlength{\tabcolsep}{1em}
	\renewcommand{\arraystretch}{1.4} %
	\caption{Results from our proposed methods on the mutual-actions subset of NTU RGB+D 120 dataset.}
	\begin{tabular}{lcc} 		\toprule
		& \multicolumn{2}{c}{\textbf{Acc Mutual Actions}} \\
		\textbf{Method} & \textbf{Cross-Subject} & \textbf{Cross-Setup} \\ \midrule
		ST-LSTM~\cite{Liu2016a}    			& 63.0\% & 66.6\% \\ 
		GCA-LSTM~\cite{Liu2017c}   			& 70.6\% & 73.7\% \\ 
		Two-stream GCA-LSTM~\cite{Liu2018}  & 73.0\% & 73.3\% \\
		FSNET~\cite{Liu2019c}      			& 61.2\% & 69.7\% \\
		ST-GCN~\cite{Yan2018a}	        	& 78.9\% & 76.1\% \\
		AS-GCN~\cite{Li2019}		    	& 82.9\% & 83.7\% \\
		\midrule
		LSTM-$IRN^{'}_{inter}$          & 74.3\% & 75.6\% \\ 
		LSTM-$IRN^{'}_{intra}$          & 73.6\% & 75.2\% \\ 
		LSTM-$IRN^{'fc1}_{inter+intra}$ & 77.7\% & 79.6\% \\ \bottomrule
	\end{tabular}
	\label{tab:ntu_v2_inter}
	\end{center}
\end{table}

\textbf{NTU RGB+D 120.}\quad 
The new version of the NTU RGB+D dataset~\cite{Liu2019} is even more challenging, as shown in Table~\ref{tab:ntu_v2_inter} containing our results together with previous work methods.
Our proposed LSTM-$IRN$ architectures can still outmatch most of the compared methods, demonstrating they are at some extent scalable with respect to the number of classes.
Moreover, as the complexity of the interaction recognition problem grows, more benefit can be obtained with the multi-relationship model LSTM-$IRN^{'fc1}_{inter+intra}$.
Claim supported by the fact that at NTU RGB+D the gain in performance obtained was approximately of 1 percentage point over LSTM-$IRN^{'}_{inter}$, meanwhile at NTU RGB+D 120 this gain was between 3 and 4 percentage points.
Our method obtains competitive results to ST-GCN~\cite{Yan2018a}, being slightly inferior at the Cross-Subject protocol, but superior at Cross-Setup.
The graph-based method AS-GCN~\cite{Li2019} have a higher scalability than ours LSTM-$IRN$, however our better performance in NTU RGB+D~\cite{Shahroudy2016} indicates the compactness of our method make it more suitable when handling less classes.

\begin{figure}[t]
	\centering
	\includegraphics[width=.49\textwidth,trim={11 11 11 13},clip]{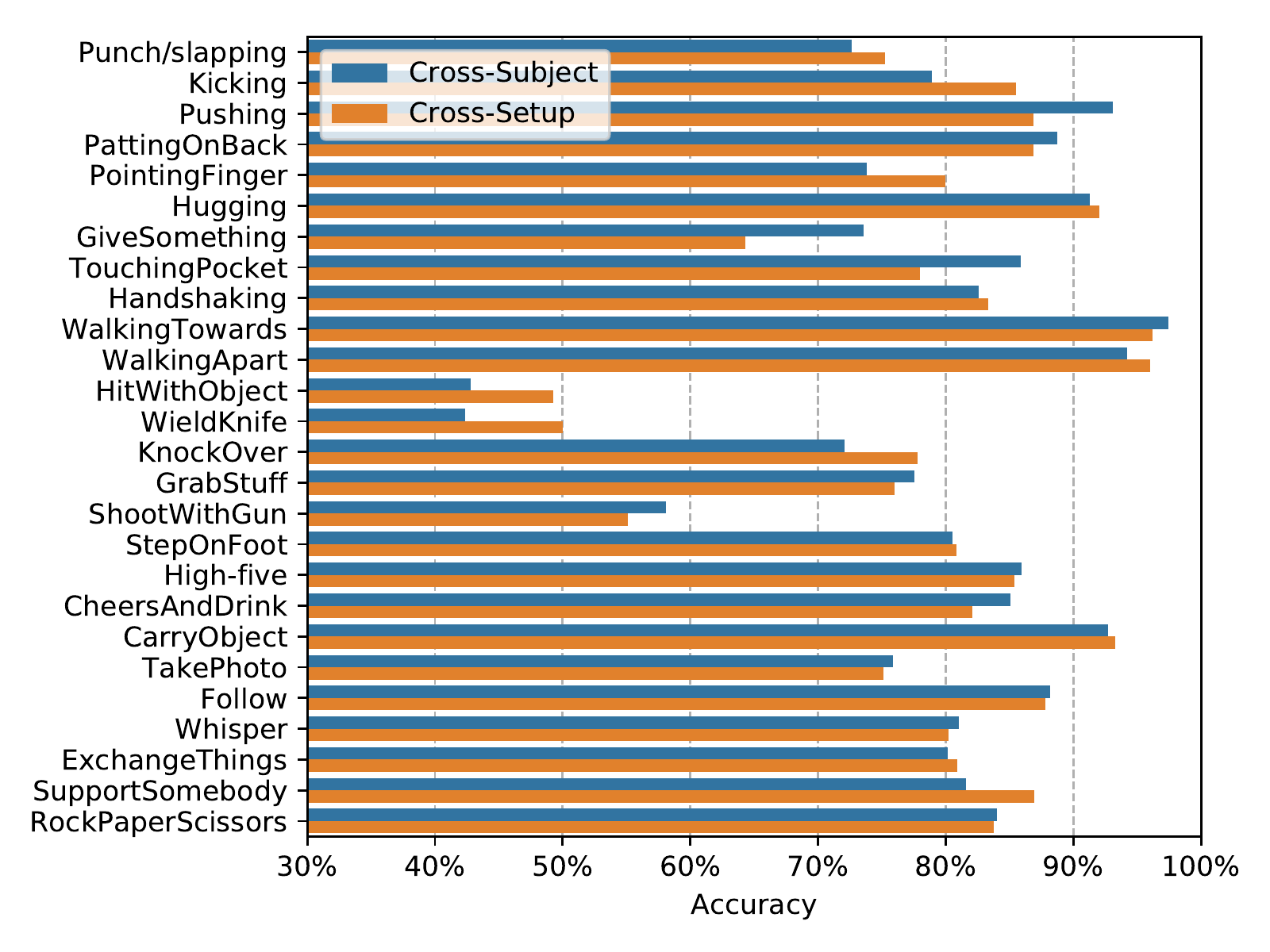}
	\caption{Performance per interaction class of method LSTM-$IRN^{'fc1}_{inter+intra}$ on the NTU RGB+D 120 dataset, for both protocols.}
	\label{fig:barplot_NTU-V2}
\end{figure}

\begin{figure*}[!t]
	\centering
	\includegraphics[width=1\textwidth, trim={15 20 21 20},clip]{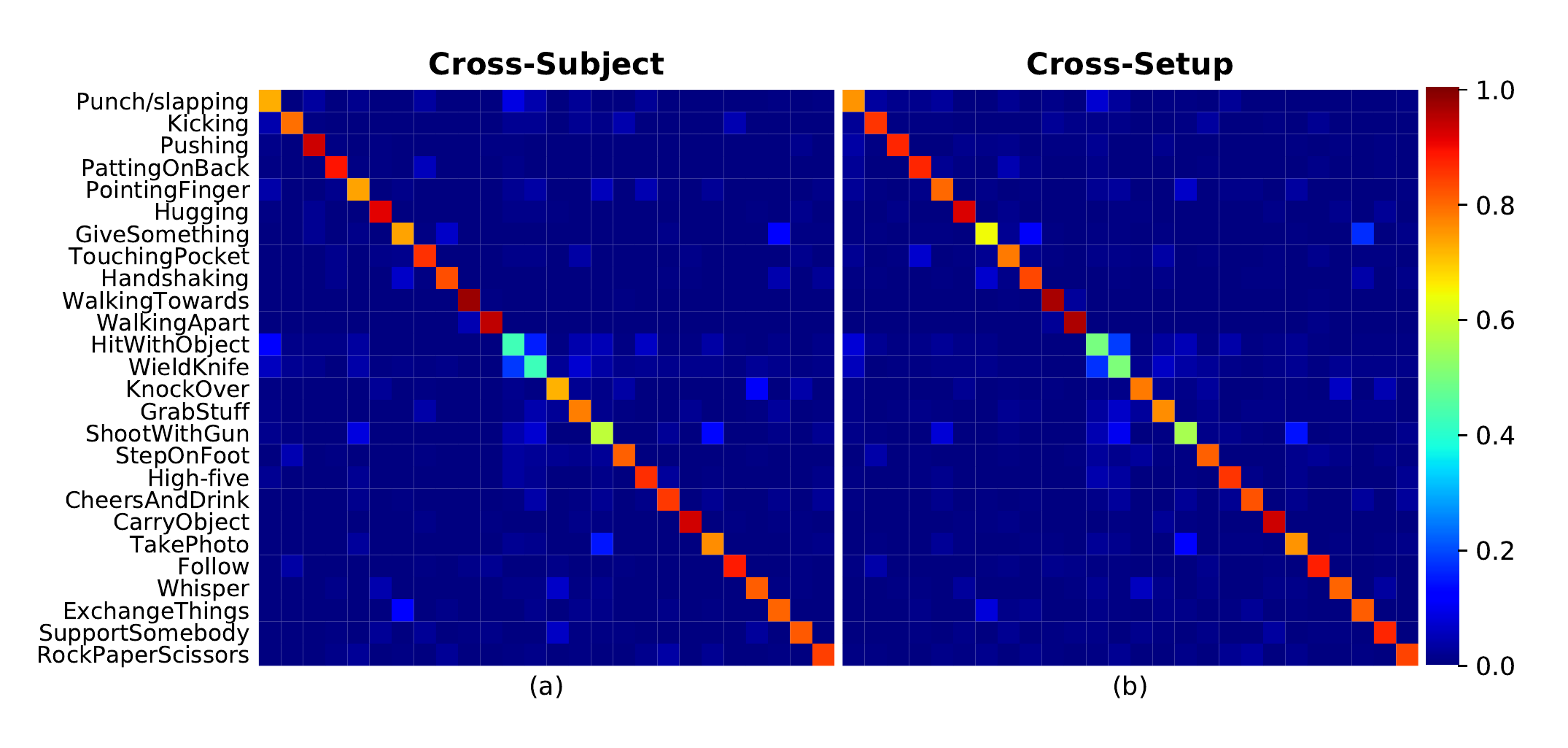}
	\caption{Confusion matrices for NTU RGB+D 120 dataset with method LSTM-$IRN^{'fc1}_{inter+intra}$ for protocols (a) Cross-Subject and (b) Cross-Setup.}
	\label{fig:confusion_matrices_ntu_v2}
\end{figure*}

A more detailed examination of the performance of LSTM-$IRN^{'fc1}_{inter+intra}$ is shown in Fig.~\ref{fig:barplot_NTU-V2}, which contains the accuracy per interaction class for both protocols. Some interactions are significantly more challenging than others, with a recognition rate much inferior than the average.
Through the confusion matrices present in Fig.~\ref{fig:confusion_matrices_ntu_v2}
it can be seen that, at least for the most severe cases (around or bellow 70\% accuracy), both protocols have confusion mostly between the same interactions.
A few notable differences are the confusion between \emph{SupportSomebody} and \emph{KnockOver} for Cross-Subject, and \emph{TouchingPocket} with \emph{PattingOnBack} for Cross-Setup.
On both protocols there is a lot of confusion between \emph{HitWithObject} and \emph{WieldKnife}, and these two with \emph{Punch/slapping}.
Also, there is some confusion between \emph{GiveSomething} with \emph{Handshaking} and \emph{ExchangeThings}, as well as between \emph{ShootWithGun} and \emph{TakePhoto}.
These confusions are reasonable, since we are using only the body parts coordinates information, and these interactions should have similar human movements (e.g. extending the hands to each other), being mainly distinguishable by the object (or its absence) each individual holds.
We believe this is a strong evidence that for more advanced interaction recognition, RGB visual information should also be used.

\section{Conclusion}
\label{sec:conclusion}

In this work, we proposed the novel Interaction Relational Network for recognition of mutual actions, through an architecture focusing at relational reasoning over the different relationships between the human joints during interactions.
Current techniques dominating the field of pose-based interaction recognition are based on CNNs/LSTMs/GCNs and consist of fixed structures determining how to relate each joint.
Therefore, given our IRN capacity to learn from the data itself how the human body parts relate to each other while two individuals interact, it stands as a highly valuable alternative to these techniques.

Our proposed method obtains state-of-the-art performance on the traditional interaction datasets SBU and UT, and it also obtains the highest performance on the mutual actions subset of the NTU RGB+D dataset, surpassing even recent graph-based methods.
Meanwhile achieving competitive results in the interactions subset of NTU RGB+D 120, what indicates it can fairly scale to more classes while being compact.
To accomplish such deed, important extensions had to be made over the original Relational Network proposition in order to tailor it for our problem and data.
Extensions such as: multiple relationships models; fusion of these multiple relationship models; domain knowledge pair-wise structured operation over the input; and sequential relational reasoning.
We hope these extensions can also be valuable to other problems, with specific data and relationship particularities that the original RN it is not capable to suitably model without it.
A more complex scenario that we believe to be a strong candidate for benefiting from our architecture and extensions, after some adaptations, is the problem of Group Activity Recognition~\cite{Ibrahim2016,Ibrahim2018}.

Furthermore, we believe our IRN can still be improved through different means.
Our relational module can benefit from other sources of data, such as high-level visual features extracted on the region around the joints coordinates, which can be appended to the current input as extra and valuable information.
Another possible improvement, would be to design a more sophisticated pair-wise structured operation. Alternatively to the fixed operations for distance and motion, which are based on the knowledge that the input data represents spatial coordinates at different frames, an architecture with trainable parameters can be designed to take leverage of the same knowledge.
A different potential place for improvement in our architecture is at the pooling stage. Instead of averaging all pairs features with the same weight, therefore giving the same importance to all pairs of joints, an attention mechanism can be applied, so as to adjust the contribution to the global descriptor depending of which joints does that relationship belong to.

\section*{Acknowledgment}

This research was carried out at the Rapid-Rich Object Search (ROSE) Lab, Nanyang Technological University (NTU), Singapore and supported by a grant from NTU’s College of Engineering (M4081746.D90).
This work was partially supported by SUTD SGP-AI grant.

\ifCLASSOPTIONcaptionsoff
  \newpage
\fi

\bibliographystyle{IEEEtran}
\bibliography{references}

\begin{thebibliography}{10}
\providecommand{\url}[1]{#1}
\csname url@samestyle\endcsname
\providecommand{\newblock}{\relax}
\providecommand{\bibinfo}[2]{#2}
\providecommand{\BIBentrySTDinterwordspacing}{\spaceskip=0pt\relax}
\providecommand{\BIBentryALTinterwordstretchfactor}{4}
\providecommand{\BIBentryALTinterwordspacing}{\spaceskip=\fontdimen2\font plus
\BIBentryALTinterwordstretchfactor\fontdimen3\font minus
  \fontdimen4\font\relax}
\providecommand{\BIBforeignlanguage}[2]{{%
\expandafter\ifx\csname l@#1\endcsname\relax
\typeout{** WARNING: IEEEtran.bst: No hyphenation pattern has been}%
\typeout{** loaded for the language `#1'. Using the pattern for}%
\typeout{** the default language instead.}%
\else
\language=\csname l@#1\endcsname
\fi
#2}}
\providecommand{\BIBdecl}{\relax}
\BIBdecl

\bibitem{Ryoo2009}
M.~S. Ryoo and J.~K. Aggarwal, ``{Spatio-Temporal Relationship Match : Video
  Structure Comparison for Recognition of Complex Human Activities},'' in
  \emph{IEEE International Conference on Computer Vision (ICCV)}, 2009, pp.
  1593--1600.

\bibitem{Rapitis2013}
M.~Raptis and L.~Sigal, ``{Poselet key-framing: A model for human activity
  recognition},'' in \emph{IEEE Conference on Computer Vision and Pattern
  Recognition (CVPR)}, 2013, pp. 2650--2657.

\bibitem{Aliakbarian2017}
M.~S. Aliakbarian, F.~S. Saleh, M.~Salzmann, B.~Fernando, L.~Petersson, and
  L.~Andersson, ``{Encouraging LSTMs to Anticipate Actions Very Early},'' in
  \emph{IEEE International Conference on Computer Vision (ICCV)}, 2017, pp.
  280--289.

\bibitem{Wang2017}
X.~Wang and Q.~Ji, ``{Hierarchical context modeling for video event
  recognition},'' \emph{IEEE Transactions on Pattern Analysis and Machine
  Intelligence (TPAMI)}, vol.~39, no.~9, pp. 1770--1782, 2017.

\bibitem{Vahdat2011}
A.~Vahdat, B.~Gao, M.~Ranjbar, and G.~Mori, ``{A discriminative key pose
  sequence model for recognizing human interactions},'' in \emph{IEEE
  International Conference on Computer Vision Workshops (ICCV Workshops)},
  2011, pp. 1729--1736.

\bibitem{Yun2012}
K.~Yun, J.~Honorio, D.~Chattopadhyay, T.~L. Berg, and D.~Samaras, ``{Two-person
  interaction detection using body-pose features and multiple instance
  learning},'' in \emph{IEEE Conference on Computer Vision and Pattern
  Recognition Workshops (CVPRW)}, 2012, pp. 28--35.

\bibitem{Ryoo2011}
M.~S. Ryoo, ``{Human activity prediction: Early recognition of ongoing
  activities from streaming video},'' in \emph{IEEE International Conference on
  Computer Vision (ICCV)}, 2011, pp. 1036--1043.

\bibitem{Zhang2012}
Y.~Zhang, X.~Liu, M.-C. Chang, W.~Ge, and T.~Chen, ``{Spatio-Temporal Phrases
  for Activity Recognition},'' in \emph{Springer European Conference on
  Computer Vision (ECCV)}, 2012, pp. 707--721.

\bibitem{Donahue2015}
J.~Donahue, L.~A. Hendricks, M.~Rohrbach, S.~Venugopalan, S.~Guadarrama,
  K.~Saenko, and T.~Darrell, ``{Long-Term Recurrent Convolutional Networks for
  Visual Recognition and Description},'' in \emph{IEEE Conference on Computer
  Vision and Pattern Recognition (CVPR)}, 2015, pp. 2625--2634.

\bibitem{Shi2018}
Y.~Shi, B.~Fernando, and R.~Hartley, ``{Action Anticipation with RBF Kernelized
  Feature Mapping RNN},'' in \emph{Springer European Conference on Computer
  Vision (ECCV)}, 2018, pp. 305--322.

\bibitem{Zhang2012a}
Z.~Zhang, ``{Microsoft Kinect Sensor and Its Effect},'' \emph{IEEE Multimedia},
  vol.~19, no.~2, pp. 4--10, 2012.

\bibitem{Cao2018}
Z.~Cao, T.~Simon, S.-E. Wei, and Y.~Sheikh, ``{OpenPose: Realtime Multi-Person
  2D Pose Estimation using Part Affinity Fields},'' \emph{arXiv preprint
  arXiv:1812.08008}, pp. 1--14, 2018.

\bibitem{Ji2014}
Y.~Ji, G.~Ye, and H.~Cheng, ``{Interactive body part contrast mining for human
  interaction recognition},'' in \emph{IEEE International Conference on
  Multimedia and Expo Workshops (ICMEW)}, 2014, pp. 1--6.

\bibitem{Ji2015}
Y.~Ji, H.~Cheng, Y.~Zheng, and H.~Li, ``{Learning contrastive feature
  distribution model for interaction recognition},'' \emph{Journal of Visual
  Communication and Image Representation}, vol.~33, pp. 340--349, 2015.

\bibitem{Li2016}
M.~Li and H.~Leung, ``{Multiview Skeletal Interaction Recognition Using Active
  Joint Interaction Graph},'' \emph{IEEE Transactions on Multimedia (TMM)},
  vol.~18, no.~11, pp. 2293--2302, 2016.

\bibitem{Wu2018a}
H.~Wu, J.~Shao, X.~Xu, Y.~Ji, F.~Shen, and H.~T. Shen, ``{Recognition and
  Detection of Two-Person Interactive Actions Using Automatically Selected
  Skeleton Features},'' \emph{IEEE Transactions on Human-Machine Systems},
  vol.~48, no.~3, pp. 304--310, 2018.

\bibitem{Perez2019a}
M.~Perez, J.~Liu, and A.~C. Kot, ``{Interaction Recognition Through Body Parts
  Relation Reasoning},'' in \emph{IAPR Asian Conference on Pattern Recognition
  (ACPR)}, 2020, pp. 268--280.

\bibitem{Du2015}
Y.~Du, W.~Wang, and L.~Wang, ``{Hierarchical recurrent neural network for
  skeleton based action recognition},'' in \emph{IEEE Conference on Computer
  Vision and Pattern Recognition (CVPR)}, 2015, pp. 1110--1118.

\bibitem{Zhu2016a}
W.~Zhu, C.~Lan, J.~Xing, W.~Zeng, Y.~Li, L.~Shen, and X.~Xie, ``{Co-Occurrence
  Feature Learning for Skeleton Based Action Recognition Using Regularized Deep
  LSTM Networks.}'' in \emph{AAAI}, vol.~2, no.~5, 2016, pp. 3697--3703.

\bibitem{Liu2016a}
J.~Liu, A.~Shahroudy, D.~Xu, and G.~Wang, ``{Spatio-Temporal LSTM with Trust
  Gates for 3D Human Action Recognition},'' in \emph{Springer European
  Conference on Computer Vision (ECCV)}, 2016, pp. 816--833.

\bibitem{Cai2016}
X.~Cai, W.~Zhou, L.~Wu, J.~Luo, and H.~Li, ``{Effective Active Skeleton
  Representation for Low Latency Human Action Recognition},'' \emph{IEEE
  Transactions on Multimedia (TMM)}, vol.~18, no.~2, pp. 141--154, 2016.

\bibitem{Yang2017}
Y.~Yang, C.~Deng, S.~Gao, W.~Liu, D.~Tao, and X.~Gao, ``{Discriminative
  Multi-instance Multitask Learning for 3D Action Recognition},'' \emph{IEEE
  Transactions on Multimedia (TMM)}, vol.~19, no.~3, pp. 519--529, 2017.

\bibitem{Liu2018}
J.~Liu, G.~Wang, L.-Y. Duan, K.~Abdiyeva, and A.~C. Kot, ``{Skeleton-Based
  Human Action Recognition with Global Context-Aware Attention LSTM
  Networks},'' \emph{IEEE Transactions on Image Processing (TIP)}, vol.~27,
  no.~4, pp. 1586--1599, 2018.

\bibitem{Ke2018a}
Q.~Ke, M.~Bennamoun, S.~An, F.~Sohel, and F.~Boussaid, ``{Learning Clip
  Representations for Skeleton-Based 3D Action Recognition},'' \emph{IEEE
  Transactions on Image Processing (TIP)}, vol.~27, no.~6, pp. 2842--2855,
  2018.

\bibitem{Ke2019}
Q.~Ke, M.~Bennamoun, H.~Rahmani, S.~An, F.~Sohel, and F.~Boussaid, ``{Learning
  Latent Global Network for Skeleton-based Action Prediction},'' \emph{IEEE
  Transactions on Image Processing (TIP)}, vol.~29, pp. 959 -- 970, 2019.

\bibitem{Liu2019d}
J.~Liu, H.~Rahmani, N.~Akhtar, and A.~Mian, ``{Learning Human Pose Models from
  Synthesized Data for Robust RGB-D Action Recognition},'' \emph{Springer
  International Journal of Computer Vision (IJCV)}, vol. 127, no.~10, pp.
  1545--1564, 2019.

\bibitem{Santoro2017}
A.~Santoro, D.~Raposo, D.~G.~T. Barrett, M.~Malinowski, R.~Pascanu,
  P.~Battaglia, and T.~Lillicrap, ``{A simple neural network module for
  relational reasoning},'' in \emph{Advances in Neural Information Processing
  Systems (NIPS)}, 2017, pp. 4967--4976.

\bibitem{Shahroudy2016}
A.~Shahroudy, J.~Liu, T.-T. Ng, and G.~Wang, ``{NTU RGB+D: A Large Scale
  Dataset for 3D Human Activity Analysis},'' in \emph{IEEE Conference on
  Computer Vision and Pattern Recognition (CVPR)}, 2016, pp. 1010--1019.

\bibitem{Liu2019}
J.~Liu, A.~Shahroudy, M.~Perez, G.~Wang, L.-Y. Duan, and A.~C. Kot, ``{NTU
  RGB+D 120: A Large-Scale Benchmark for 3D Human Activity Understanding},''
  \emph{IEEE Transactions on Pattern Analysis and Machine Intelligence
  (TPAMI)}, vol.~42, no.~10, pp. 2684--2701, 2019.

\bibitem{Wang2013}
H.~Wang and C.~Schmid, ``{Action Recognition with Improved Trajectories},'' in
  \emph{IEEE International Conference on Computer Vision (ICCV)}, 2013, pp.
  3551--3558.

\bibitem{Simonyan2014}
K.~Simonyan and A.~Zisserman, ``{Two-Stream Convolutional Networks for Action
  Recognition in Videos},'' in \emph{Advances in Neural Information Processing
  Systems (NIPS)}, 2014, pp. 568--576.

\bibitem{Wu2016}
L.~Wang, Y.~Xiong, Z.~Wang, Y.~Qiao, D.~Lin, X.~Tang, and L.~{Van Gool},
  ``{Temporal Segment Networks: Towards Good Practices for Deep Action
  Recognition},'' in \emph{Springer European Conference on Computer Vision
  (ECCV)}, 2016, pp. 20--36.

\bibitem{Shi2017a}
Y.~Shi, Y.~Tian, Y.~Wang, and T.~Huang, ``{Sequential Deep Trajectory
  Descriptor for Action Recognition With Three-Stream CNN},'' \emph{IEEE
  Transactions on Multimedia (TMM)}, vol.~19, no.~7, pp. 1510--1520, 2017.

\bibitem{Ke2018}
Q.~Ke, M.~Bennamoun, S.~An, F.~Sohel, and F.~Boussaid, ``{Leveraging Structural
  Context Models and Ranking Score Fusion for Human Interaction Prediction},''
  \emph{IEEE Transactions on Multimedia (TMM)}, vol.~20, no.~7, pp. 1712--1723,
  2018.

\bibitem{Chowdhury2018}
M.~I.~H. Chowdhury, K.~Nguyen, S.~Sridharan, and C.~Fookes, ``{Hierarchical
  Relational Attention for Video Question Answering},'' in \emph{IEEE
  International Conference on Image Processing (ICIP)}, 2018, pp. 599--603.

\bibitem{Park2018}
S.~Park and N.~Kwak, ``{3D Human Pose Estimation with Relational Networks},''
  in \emph{British Machine Vision Conference (BMVC)}, 2018, pp. 1--13.

\bibitem{Ibrahim2018}
M.~S. Ibrahim and G.~Mori, ``{Hierarchical Relational Networks for Group
  Activity Recognition and Retrieval},'' in \emph{Springer European Conference
  on Computer Vision (ECCV)}, 2018, pp. 721--736.

\bibitem{Yu2016}
F.~Yu and V.~Koltun, ``{Multi-Scale Context Aggregation by Dilated
  Convolutions},'' in \emph{International Conference on Learning
  Representations (ICLR)}, 2016, pp. 1--13.

\bibitem{Li2015}
W.~Li, L.~Wen, M.~C. Chuah, and S.~Lyu, ``{Category-Blind Human Action
  Recognition: A Practical Recognition System},'' in \emph{IEEE International
  Conference on Computer Vision (ICCV)}, 2015, pp. 4444--4452.

\bibitem{Zhang2017}
P.~Zhang, C.~Lan, J.~Xing, W.~Zeng, J.~Xue, and N.~Zheng, ``{View Adaptive
  Recurrent Neural Networks for High Performance Human Action Recognition from
  Skeleton Data},'' in \emph{IEEE International Conference on Computer Vision
  (ICCV)}, 2017, pp. 2136--2145.

\bibitem{Kong2016}
Y.~Kong and Y.~Fu, ``{Close human interaction recognition using patch-aware
  models},'' \emph{IEEE Transactions on Image Processing (TIP)}, vol.~25,
  no.~1, pp. 167--178, 2016.

\bibitem{Shu2018}
X.~Shu, J.~Tang, G.-J. Qi, W.~Liu, and J.~Yang, ``{Hierarchical Long Short-Term
  Concurrent Memory for Human Interaction Recognition},'' \emph{IEEE
  Transactions on Pattern Analysis and Machine Intelligence (TPAMI)}, vol.
  1811.00270, pp. 1--1, 2019.

\bibitem{Liu2017c}
J.~Liu, G.~Wang, P.~Hu, L.-Y. Duan, and A.~C. Kot, ``{Global Context-Aware
  Attention LSTM Networks for 3D Action Recognition},'' in \emph{IEEE
  Conference on Computer Vision and Pattern Recognition (CVPR)}, 2017, pp.
  1647--1656.

\bibitem{Liu2019c}
J.~Liu, A.~Shahroudy, G.~Wang, L.-Y. Duan, and A.~C. Kot, ``{Skeleton-Based
  Online Action Prediction Using Scale Selection Network},'' \emph{IEEE
  Transactions on Pattern Analysis and Machine Intelligence (TPAMI)}, vol.~42,
  no.~6, pp. 1453--1467, 2019.

\bibitem{Yan2018a}
S.~Yan, Y.~Xiong, and D.~Lin, ``{Spatial Temporal Graph Convolutional Networks
  for Skeleton-Based Action Recognition},'' in \emph{AAAI Conference on
  Artificial Intelligence}, 2018.

\bibitem{Li2019}
M.~Li, S.~Chen, X.~Chen, Y.~Zhang, Y.~Wang, and Q.~Tian, ``{Actional-Structural
  Graph Convolutional Networks for Skeleton-Based Action Recognition},'' in
  \emph{IEEE Conference on Computer Vision and Pattern Recognition (CVPR)},
  2019, pp. 3590--3598.

\bibitem{Ibrahim2016}
M.~S. Ibrahim, S.~Muralidharan, Z.~Deng, A.~Vahdat, and G.~Mori, ``{A
  Hierarchical Deep Temporal Model for Group Activity Recognition},'' in
  \emph{IEEE Conference on Computer Vision and Pattern Recognition (CVPR)},
  2016, pp. 1971--1980.

\end{thebibliography}

\begin{IEEEbiography}[{\includegraphics[width=1in,height=1.25in,clip,keepaspectratio]{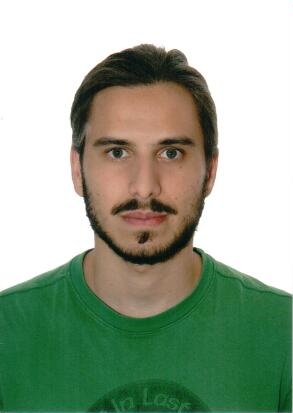}}]{Mauricio Perez}
is currently pursuing the Ph.D. degree with the School of Electrical and Electronic Engineering, Nanyang Technological University, Singapore. He received his M.Sc. degree from University of Campinas in 2016, and his B.Sc. degree from Federal University of São Carlos in 2012, both in Brazil.

His research interests include Deep Learning, Video Analysis, Sensitive Media Detection, Medical Imaging and Computer Vision.
\end{IEEEbiography}

\begin{IEEEbiography}[{\includegraphics[width=1in,height=1.25in,clip,keepaspectratio]{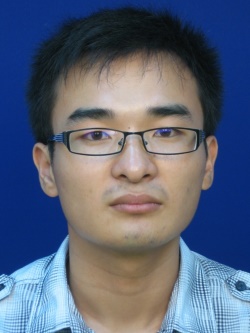}}]{Jun Liu}
is an Assistant Professor at Singapore University of Technology and Design. He 
obtained the B.Eng. degree from Central South University, China, in 2011, the M.Sc. degree from Fudan University, China, in 2014, and the Ph.D. degree with the School of Electrical and Electronic Engineering, Nanyang Technological University, Singapore, in 2019.

His research interests include video analysis, human action recognition, pose estimation, and deep learning.
\end{IEEEbiography}

\vfill

\newpage

\begin{IEEEbiography}[{\includegraphics[width=1in,height=1.25in,clip,keepaspectratio]{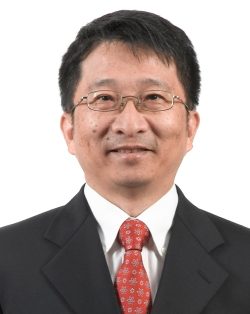}}]{Alex C. Kot}
(F'06) has been with Nanyang Technological University, Singapore, since 1991. He headed the Division of Information Engineering with the School of Electrical and Electronic Engineering for eight years and served as an Associate Chair (Research). He was the Vice Dean (Research) with the School of Electrical and Electronic Engineering and the Associate Dean for the College of Engineering for eight years. He is currently a Professor with the School of Electrical and Electronic Engineering and the Director of the Rapid-Rich Object Search Lab. He has published extensively in the areas of signal processing for communication, biometrics, image forensics, information security, and computer vision. He was a recipient of the Best Teacher of the Year Award and co-authored several best paper awards, including for ICPR, WIFS, and IWDW. He was awarded as the IEEE Distinguished Lecturer of the Signal Processing Society.

He served as the Associate Editor for the IEEE Transactions onImage Processing (T-IP), IEEE Transactions on Signal Processing (T-SP), IEEE Transactions on Multimedia (T-MM), IEEE Signal Processing Magazine (SPM), IEEE Transactions on Circuits and Systems for Video Technology (T-CSVT), and IEEE Transactions on Information Forensics and Security (T-IFS). He has served the IEEE Signal Processing Society in various capacities, such as the General Co-Chair at the 2004 IEEE International Conference on Image Processing and the Vice President of the IEEE Signal Processing Society. He is a fellow of the Academy of Engineering, Singapore, a Fellow of IEEE, and a Fellow of IES.
\end{IEEEbiography}

\vfill

\end{document}